\pdfoutput=1

\documentclass[11pt]{article}

\usepackage[]{acl}

\usepackage{times}
\usepackage{latexsym}

\usepackage[T1]{fontenc}

\usepackage[utf8]{inputenc}

\usepackage{microtype}

\usepackage{booktabs,subcaption} 
\usepackage{latexsym}
\usepackage{amsmath}
\usepackage{xspace}
\usepackage{amssymb}
\usepackage{graphicx}
\usepackage{multirow}
\usepackage{enumitem}

\DeclareMathOperator{\softmax}{softmax}
\DeclareMathOperator{\crossentropy}{cross-entropy}

\DeclareMathOperator{\LSTM}{LSTM}
\DeclareMathOperator{\RNN}{RNN}
\DeclareMathOperator{\CAP}{Cap}

\newcommand{\ie}{\emph{i.e.,}\xspace}
\newcommand{\eg}{\emph{e.g.,}\xspace}

\newcommand{\paratitle}[1]{\vspace{0.8ex}\noindent \textbf{#1}}
\newcommand{\datasetname}{Customer Service Dialog\xspace}

\title{Chat-Capsule: A Hierarchical Capsule for Dialog-level Emotion Analysis}

\author{Yequan Wang$^{1}$, Xuying Meng$^{2}$, Yiyi Liu$^{2}$, Aixin Sun$^{3}$, Yao Wang$^{4}$, Yinhe Zheng$^{5, 6}$, Minlie Huang$^{6}$\\
    $^{1}$Beijing Academy of Artificial Intelligence, Beijing, China\\
    $^{2}$Institute of Computing Technology, Chinese Academy of Sciences, Beijing, China\\
    $^{3}$School of Computer Science and Engineering, Nanyang Technological University, Singapore \\
    $^{4}$Amazon Web Services \\
    $^{5}$Lingxin AI, Beijing, China \\
    $^{6}$Department of Computer Science and Technology, Tsinghua University, Beijing, China \\
\tt tshwangyequan@gmail.com, \tt \{mengxuying, liuyiyi\}@ict.ac.cn}

\begin{document}
\maketitle
\begin{abstract}
Many  studies on dialog emotion analysis focus on utterance-level emotion only. These models hence are not optimized for dialog-level emotion detection, \ie to predict the emotion category of a dialog as a whole. More importantly, these models cannot benefit from the context provided by the whole dialog.  In real-world applications,  annotations to dialog could fine-grained, including both utterance-level tags (\eg speaker type, intent category, and emotion category), and dialog-level tags (\eg user satisfaction, and emotion curve category). 
In this paper, we propose a Context-based Hierarchical Attention Capsule~(Chat-Capsule) model, which models both utterance-level and dialog-level emotions and their interrelations. On a dialog dataset collected from customer support of an e-commerce platform, our model is also able to predict user satisfaction and emotion curve category.
Emotion curve refers to the change of emotions along the development of a conversation. 
Experiments show that the proposed Chat-Capsule outperform state-of-the-art baselines on both benchmark dataset and proprietary dataset.
Source code will be released upon acceptance.

\end{abstract}

\section{Introduction}
\label{sec:intro}

Emotion analysis, a fundamental task in natural language processing, aims to analyze people's emotions, attributes, and opinions based on written language~\cite{liu2012sentiment,pang2008opinion}. 
Existing studies mainly focus on analyzing opinions at document-level~\cite{tang2015document,wang2018sentiment} or aspect-level~\cite{wang2016attention, tang2015target}. 
Recently, emotion analysis in conversations has received much attention from both academic and industry for its widespread applications~\cite{DBLP:conf/aaai/MajumderPHMGC19,DBLP:conf/emnlp/GhosalMPCG19}. 

\begin{figure}
    \centering
    \includegraphics[width=1.029\columnwidth]{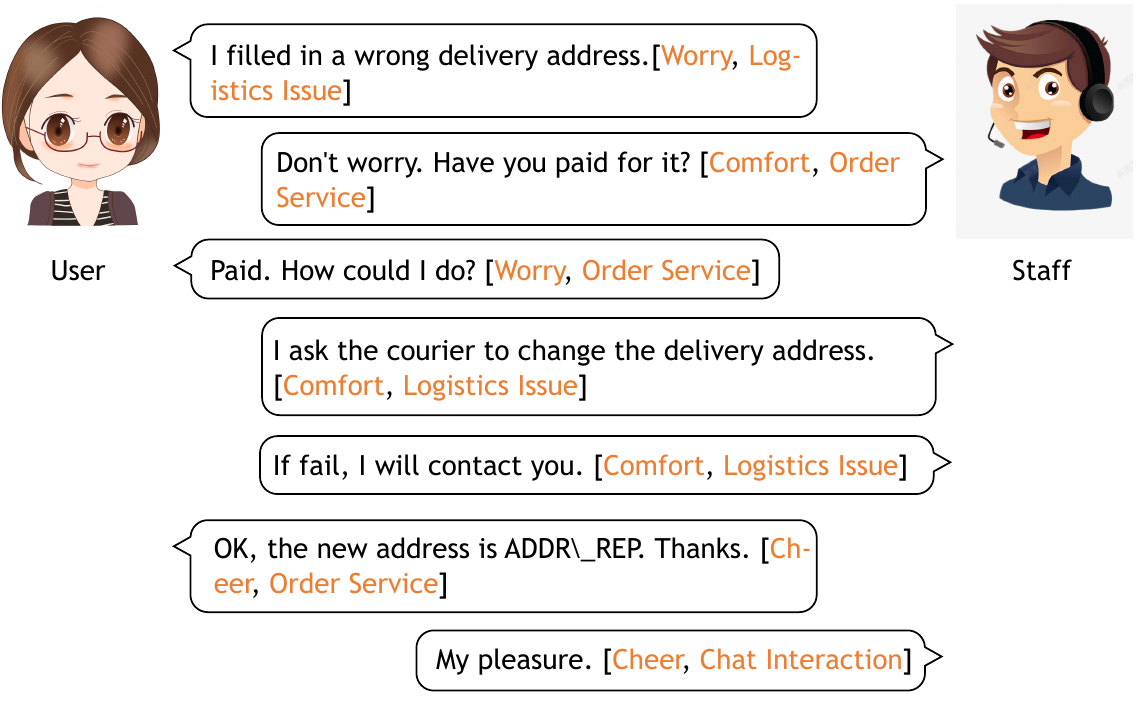}
    \caption{A simplified example of customer service dialog between user and staff. Emotion and intent of each utterance are indicated in the square bracket after it.}
    \label{fig:example}
\end{figure}

Many studies on emotion analysis in conversations focus on utterance-level emotion, but not considering the dialog as a whole. In multiple turn conversations, the participants play different roles. Figure~\ref{fig:example} shows a simplified example dialog. Shown in the example, a speaker's emotion could be affected by various contexts such as intents or responses from other speakers. Hence, emotion of all participants in a dialog could be very dynamic. In other words, it is important to take into consideration both utterance-level and dialog-level information. The change of emotions in the example dialog also reflects the quality of service from the service provider point of view.

\paratitle{The Research Formulation.}
In dialog-level emotion analysis, we have a predefined set of emotions $\mathcal{E} =\{o_1, o_2, \dots, o_{N_1}\} $, a set of intents $\mathcal{I} = \{a_1, a_2, \dots, a_{N_2}\}$, and a set of speaker types $\{m_1, m_2, \dots, m_{N_3}\}$. Given a dialog with annotated utterances $u_i$'s, denoted by $D=\{<u_1, a_1, m_1>, <u_2, a_2, m_2>, \dots, <u_L, a_L, m_L>\}$, the task is to predict  utterance-level emotion polarity $o_i$ of utterance $u_i$, and dialog-level emotions, including user satisfaction $s_d$, and emotion curve category $s_c$.
Similar to the example in Figure~\ref{fig:example},  participants in conversations are allowed to generate continuous utterances. The corresponding emotion and intent categories of utterance may be change.

Recent studies on dialog-level emotion analysis have three main branches: (i) process constituent utterances of dialogue as a sequence, such as DialogueRNN~\cite{DBLP:conf/aaai/MajumderPHMGC19}; (ii) process the multiple turns in conversation as a graph, then use graph neural network to utilize the global opinions, such as DialogueGCN~\cite{DBLP:conf/emnlp/GhosalMPCG19}; (iii) joint  emotion detection and intent category classification, known as DCR-Net~\cite{DBLP:conf/aaai/QinCLN020}.
DialogueRNN is based on recurrent neural network and keeps track of the individual party states throughout the conversation, and uses this information for emotion classification.  Stated in~\citet{DBLP:conf/emnlp/GhosalMPCG19}, RNN-based models are not effective in various tasks, including multiple turns emotion analysis in dialogs.
To remedy,  DialogueRNN employs attention that pools information from entirety or part of the conversation per target utterance. However, the attention mechanism does not consider speaker information of the utterances and their relative positions to the target utterance. 
To mitigate this problem, DialogueGCN based on GCN~\cite{DBLP:conf/esws/SchlichtkrullKB18,DBLP:conf/nips/DefferrardBV16} is proposed to leverage these two factors, by modeling conversation using a directed graph.
DialogueGCN is capable of leveraging dependency of the interlocutors to model conversational context for emotion classification.

To consider the correlation between dialog act recognition and emotion classification, joint models are proposed to solve these two tasks simultaneously in a unified framework.
DCR-Net~\cite{DBLP:conf/aaai/QinCLN020} is proposed to explicitly consider the cross-impact, and model the interaction between dialog act recognition and emotion classification with a co-interactive relation layer.
It is worth noting that a major limitation of DialogueGCN and DCR-Net is that, they require all utterances to predict utterance-level emotion.

Inspired by the great performance of capsule models, we propose Context-based Hierarchical Attention Capsule~(Chat-Capsule) for dialog-level emotion analysis. Chat-Capsule model includes  utterance-level capsule and dialog-level capsule, which is the meaning of hierarchical structure.
Recall that speaker type, intent of utterance, and contextual information all affect emotions in dialogs. Hence, the proposed Chat-Capsule models all these factors.  Specifically,  the rectifier module in utterance capsule utilizes the correlation between utterance and utterance's profile (\eg speaker type and intent). The attention guided by speaker type and intent in utterance capsule strengthen the representation learning. The feedback module in utterance capsule  allows a two-way flow of information between hierarchical capsules. Lastly, the context-based attention in dialog capsule captures contextual information. Our main contributions are as follows:
\begin{itemize}
    \item We propose Context-based Hierarchical Attention Capsule~(Chat-Capsule) for dialog-level emotion analysis. Our model  utilizes contextual correlation and the crucial factors of dialog such as speaker type and intent.
    \item Chat-Capsule does not rely on all utterances of a dialog. It only needs the current and previous utterances to predict the current utterance's emotion. 
    \item Experiments show that our approach improves the performance of three tasks, including utterance-level emotion classification, user satisfaction classification, and emotion curve detection.
\end{itemize}

\section{Related Work}
\label{sec:relatedwork}

Emotion analysis has received much attention in natural language processing~\cite{colnerivc2018emotion,DBLP:journals/corr/abs-1803-06397}.
Ekman finds correlation between emotion and facial cues~\cite{ekman1993facial}.
Datcu and Rothkrantz fuse acoustic information with visual cues for emotion recognition~\cite{datcu2014semantic}.
There are three main approaches to recognize emotions in conversation, including sequence-based structure, graph-based structure, and joint structure. 

The mainstream approach is based on sequence structure. Recurrent Neural Network (RNN)~\cite{mikolov2012statistical,tai2015improved} is a popular basic model. 
\citet{DBLP:conf/emnlp/ZadehCPCM17} uses RNN for multimodal emotion recognition. 
To model  emotional dynamics, a party state and global state based recurrent model known as DialogueRNN is proposed to recognize emotion in conversations~\cite{DBLP:conf/aaai/MajumderPHMGC19}. 

Another popular mainstream approach to recognize emotion is based on graph structure. 
Graph neural networks~\cite{gori2005new} have also been a popular choice recently and have been applied to sentiment analysis~\cite{DBLP:conf/ijcnn/ChenHJG19}.
For instance, to extend DialogueRNN to consider speaker information of the utterances and the relative positions of other utterances from the target utterance, \citet{DBLP:conf/emnlp/GhosalMPCG19} proposes DialogueGCN based on GCN~\cite{DBLP:conf/esws/SchlichtkrullKB18,DBLP:conf/nips/DefferrardBV16} to leverage these two factors by modelling conversation using a directed graph.

Considering the correlation between dialog act recognition and emotion classification, joint models are proposed to solve these two tasks simultaneously, in a unified framework.
Emotion of utterance is strongly correlated with aspect~\cite{wang2016attention}.
Intent, also known as dialog act type~\cite{stolcke1998dialog}, is also important for classifying emotion category. 
Interestingly,  speaker type and intent of utterance play similar roles in dialog-level emotion analysis, similar to aspect for aspect-level sentiment analysis.
Considering different intents, emotion has a different distribution.
DCR-Net~\cite{DBLP:conf/aaai/QinCLN020}  explicitly models the correlation between dialog act recognition and emotion classification 
by sharing parameters.

\section{Chat-Capsule Model}
\label{sec:model}

In this work, we follow the high-level concept of capsule in RNN-Capsule~\cite{wang2018sentiment}. RNN-Capsule is designed to classify sentiment~(\eg positive, neutral, or negative) of the given text.
As Chat-Capsule model is based on capsule, so we give a preliminary of RNN-Capsule before detailing the structure of our model. We refer readers to~\citet{wang2018sentiment} for full details of RNN-Capsule.

\begin{figure*}
    \centering
    \includegraphics[width=2.1\columnwidth]{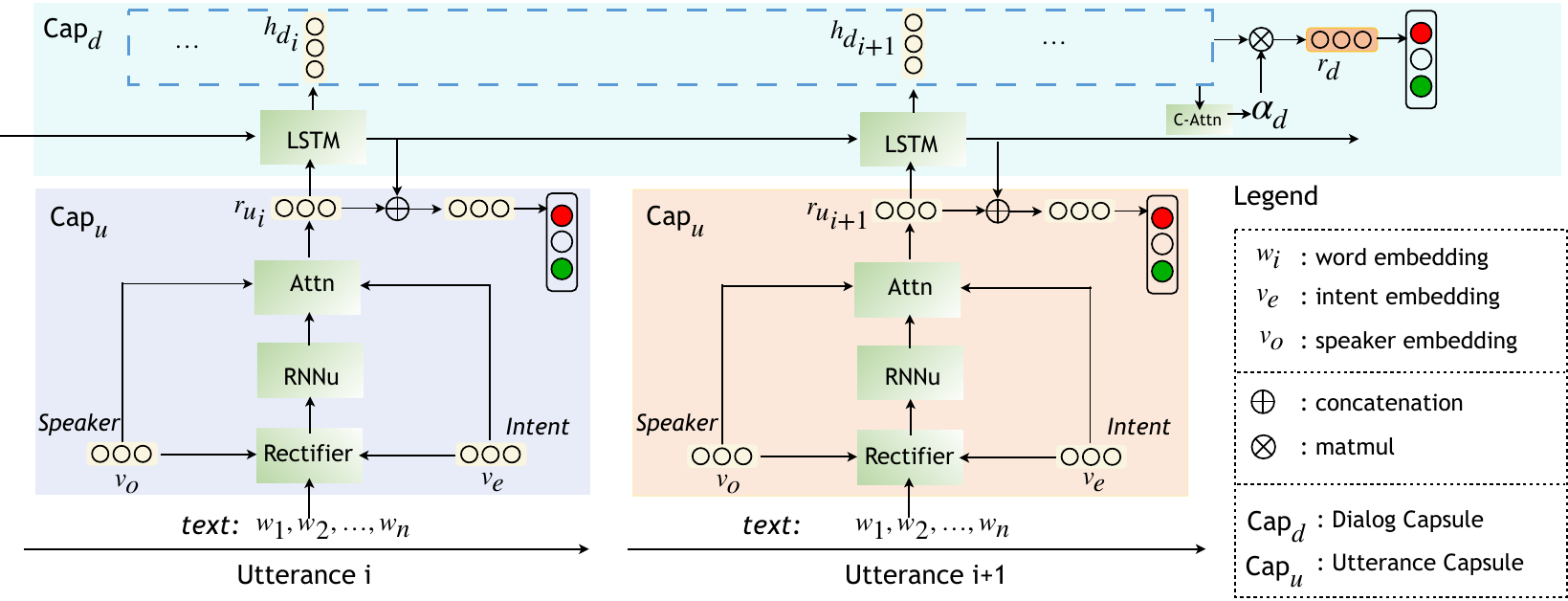}
    \caption{The architecture of the Chat-Capsule model. 
    Utterances (\ie messages from staff or user) are shown on the bottom. 
    Utterance capsule $\CAP_u$ constitutes utterance encoder, which include rectifier module, RNN, and attention mechanism. Dialog Capsule $\CAP_d$ composed by $\LSTM$ and Attention serves as dialog encoder. Both $\CAP_u$ and $\CAP_d$ consider speaker embedding $v_o$, and intent embedding $v_e$. 
    }
    \label{fig:archi}
    \vspace{1ex}
\end{figure*}

\subsection{Preliminary: RNN-Capsule}

\paratitle{Recurrent Neural Network.}
As the name suggests, RNN-Capsule is based on RNN. A recurrent neural network~(RNN) is able to exhibit dynamic temporal behavior for a sequence through connections between units, \eg LSTM, GRU, or their variants.
Briefly speaking, in an RNN realized by LSTM,  the hidden states $h_t$ and memory cell $c_t$ in LSTM is a function of the previous $h_{t-1}$ and $c_{t-1}$, and the input vector $x_t$, or formally as follows:
\begin{equation}
    c_t, h_t = \LSTM(c_{t-1}, h_{t-1}, x_t),
\end{equation}
where hidden state $h_t$ represents the representation of position $t$ while encoding the preceding contexts of the position. More details about LSTM are given in~\citet{hochreiter1997long}.

\paratitle{RNN-Capsule.} RNN-Capsule is proposed to recognize sentiment (\eg positive, negative) of the given text. The input text is encoded by LSTM~\cite{hochreiter1997long} and the hidden vector representations are input to all capsules. One capsule is built for one sentiment category, and each capsule is built with an attribute, a state, and three modules. The dedicated sentiment category is reflected in the attribute of the corresponding capsule. There are three modules in capsules including (i) representation module for building capsule representation by attention mechanism, (ii) probability module for predicting the capsule's state probability based on its representation, and (iii) reconstruction module for rebuilding the representation of the input text. A capsule's state is `active' if the output of its probability module is the largest among all capsules, and `inactive' otherwise.

There are two training objectives in RNN-Capsule. The first is to maximize the state probability of the capsule corresponding to the groundtruth sentiment, and  to minimize the state probabilities of other capsule(s) simultaneously. The second is  to minimize the distance between the input representation and the reconstruction representation of the capsule corresponding to the ground truth, and to maximize such distances for other capsule(s).

RNN-Capsule is not designed for dialog-level emotion analysis, and each capsule in RNN-Capsule corresponds to one sentiment category. A remarkable feature of dialog is  the hierarchical text structure of conversation. Dialog-level emotion analysis hence shall take into consideration not only the utterance-level emotion, but also the dialog-level opinion. Accordingly, how to design structure to benefit from both subtasks are crucial. That is, well-designed communication component for utterance-level and dialog-level subtasks will both benefit the overall objective. 

\subsection{Structure of Chat-Capsule Model}

The architecture of the Chat-Capsule is depicted in Figure~\ref{fig:archi}. 
The general dialog is shown in Figure~\ref{fig:example}, where messages from user and staff are shown on the left and right hand side respectively. 
For simplicity, only the $i$-th and $i+1$-th utterances are shown in the structure.
We use hierarchical capsules including utterance capsule and dialog capsule to model a dialog. Utterance Capsule $\CAP_u$ encodes an utterance held by user or staff. Dialog Capsule $\CAP_d$ utilizes the utterance representations and their contexts.
Recall that emotion is affected by speaker type and intent. Inspired by~\citet{wang2016attention,wang2018sentiment}, we propose speaker embeddings $v_o$ and intent embeddings $v_e$.
Speaker embeddings and intent embeddings are the very first attempt to utilize the correlation between emotion and the speaker type, and also intent of utterance. We apply them to both utterance-level and dialog-level encoders.
Context-based attention in dialog capsule is able to utilize the contextual correlation to predict  user satisfaction and emotion curve type, which is enlightened by~\citet{bahdanau2014neural,rocktaschel2015reasoning,wang2016attention}.
Moreover, the feedback module allows backward and forward information flow to further improve the performance and capability of our model.

\paratitle{Utterance-level Capsule.} An utterance-level capsule encodes each utterance to a vector representation and classifies the emotion of the input utterance. The $\CAP_d$ in Figure~\ref{fig:archi} shows the detailed architecture of the utterance capsule. There are three components: rectifier, utterance encoder $\RNN_u$, and attention.

Given an utterance with $N$ words, $\{w_1, w_2, \dots, w_N\}$, rectifier uses similarity to scale the speaker embedding $v_o$ and intent embedding $v_e$,
\begin{align}
    v'_{o} & =  v_o * \cos(w_i, v_o) \\
    v'_{e} & = v_e * \cos(w_i, v_e) \\
    w'_i & = [w_i, v'_o, v'_e]
\end{align}
Here, the word vectors $w_i$ of the input utterance  can be obtained from Glove~\cite{pennington2014glove}, word2vec~\cite{mikolov2013distributed} or other embeddings~\cite{song2018directional}. The input of $\RNN_u$ is the concatenation of word vector and the scaled intent embedding $v_e$ and speaker embedding $v_o$.
The operator brackets represent concatenation. Then the $\RNN_u$ in this capsule encodes $W=\{w'_1, w'_2, \dots, w'_N\}$ and outputs hidden representations $H_u$, or formally:
\begin{equation}
    H_u = \RNN_u(W)
\end{equation}

After getting $H_u$, utterance attention is adopted to attend the important part of utterance.
We compute the utterance representation $r_u$ using attention:
\begin{align}
    M_u & = \text{ReLU}(W_{u_1}H_u + W_{u_2}v_o\otimes N \\ \nonumber
    & \quad \quad \quad \quad \quad \quad \quad \quad + W_{u_3}v_e\otimes N)\\
    \alpha_{u} & = \softmax(w^T_u M_u) \\
    r_u & = H_u \alpha_{u}^T
\end{align}
Here, $W_{u_1}$, $W_{u_2}$, $W_{u_3}$ and $w_u$ are the parameters of this attention.
$v_o \otimes N $ is the operator that repeatedly concatenates $v_o$ for $N$ times.
Lastly, the utterance representation vector $r_u$ is a weighted summation over all the positions using  attention importance scores as weights.

The utterance-level capsule by design has the ability to communicate with dialog-level capsule. Dialog capsule will feedback the current information $h_{d_{i}}$ of all the utterances before $i$ to the $i$-th $\CAP_u$. With this information, the utterance capsule could predict the emotion of utterance at a high level.
\begin{equation}
    \mathcal{P}_i= \softmax(W_{p} [r_{u_i}, h_{d_{i}}] + b_{p}),
\end{equation}
where $W_p$ and $b_p$ are the parameters for the emotion distribution
module of the current utterance-level capsule.

\paratitle{Dialog-level Capsule.} $\RNN_d$ encodes the utterance representations received from utterance-level capsules. The input of $\RNN_d$ is concatenated with $v_o$ and $v_e$, which is capable of utilizing the speaker type and intent information. For simplicity, the concatenation operation is not shown in Figure~\ref{fig:archi}.
As aforementioned, the emotion distribution changes with speaker type. Hence, speaker embeddings are capable of utilizing the correlation effectively.
Briefly speaking, given utterance representations from utterance-level encoders, we obtain a high-level dialog representation with $M$ utterances through
\begin{equation}
    H_d = \RNN_d(X, v_o, v_e),
\end{equation}
where $\RNN_d$ is the dialog encoder. $X=\{r_{u_1}, r_{u_2}, \dots, r_{u_M}\}$, and $r_{u_i}$ denotes the utterance representation of the $i$-th utterance encoded by utterance-level capsule.

Again, the emotion of a dialog is affected by the trend of emotion change. How to model the change tendency is a key challenge in this task. Hence, we design context-based attention to utilize the contextual correlation in a dialog.
After getting the hidden representations $H_d$ from dialog-level encoder $\RNN_d$, we compute the dialog representation $r_d$ using context-based attention through
\begin{align}
    e_d & = w_d H^T_d\\
    \alpha_{d} & = \softmax(e_d) \\
    r_d & = H_d \alpha_{d}^T
\end{align}
In the above formula, $H_d$ denotes the hidden representations of $\RNN_d$. $w_d$ is the parameter of this attention layer. 
The importance score is $\alpha_d$.

Note that, the dialog representation $r_d$ is obtained from the context-based attention which is a high-level encoding of the whole dialog. We observe that the attention is able to improve the model's performance in experiments.

\subsection{Learning Objective}

The proposed Chat-Capsule model has two learning objectives.
One is to minimize the cross-entropy of emotion distribution at utterance-level; the other is to minimize the cross-entropy of emotion distribution at dialog-level.

\paratitle{Utterance-level Emotion Objective.} Each utterance held by either user or staff is encoded by the utterance-level encoder. To minimize the cross entropy of emotion at utterance-level, the emotion classification objective $J$ can be formulated as:
\begin{equation}
    J(\theta) = \sum_i\crossentropy(y^i_{u}, \mathcal{P}_i),
\end{equation}
where $\mathcal{P}_i$ denotes the emotion probability distribution of utterance $i$ in dialog. The groundtruth of emotion category is $y^i_{u}$ at utterance-level.

\paratitle{Dialog-level opinion Objective.}  Emotion analysis at utterance-level and dialog-level has a strong correlation.  Unavoidably, the opinion of a dialog is heavily affected by the emotions of its utterances. Meanwhile, the utterance-level opinion also depends on the emotion tendency in the dialog. Similarly, the dialog-level opinion classification objective can be formulated as:
\begin{equation}
    U(\theta) = \sum_j\crossentropy(y^j_{d}, \mathcal{P}'_j),
\end{equation}
where $\mathcal{P}'_j$ is the opinion probability distribution of dialog $j$. The groundtruth of user satisfaction is $y^j_{d}$ at dialog-level. The computing method of opinion curve is exactly the same as user satisfaction.

Considering the two objectives, we get our final objective function $L$ by adding $J$ and $U$:
\begin{equation}
    L(\theta) = J(\theta) + U(\theta),
\end{equation}
where $\theta$ is the parameter set of the model.

In the proposed Chat-Capsule model, both utterance-level and dialog-level capsules utilize the speaker type and intent of each utterance. As the emotion of dialog is largely affected by the contextual information, we adopt  two-level encoders and context-based attention to model the hierarchical structure of dialog, and utilize the information in the trend of opinion change. All of these carefully designed components work together to improve the model's capability and robustness.

\section{Experiment}
\label{sec:experiment}

\subsection{Dataset}

We evaluate Chat-Capsule model for dialog-level emotion analysis two datasets. One is  a benchmark dataset named DailyDialog~\cite{DBLP:conf/ijcnlp/LiSSLCN17}, and the other is our proprietary dataset, named Customer Service Dialog. DailyDialog dataset has been widely used in dialog-level emotion analysis evaluation, which enables us to benchmark our result against the published results. Specifically, we evaluate Chat-Capsule for both utterance-level and dialog-level emotion analysis, against baselines.
We do not use IEMOCAP~\cite{DBLP:journals/lre/BussoBLKMKCLN08} due to its small size, having only 151 dialogs.

\paratitle{DailyDialog.} 
DailyDialog~\cite{DBLP:conf/ijcnlp/LiSSLCN17} adopts daily conversations written by humans in English. we adopt the standard split from the original dataset~\cite{DBLP:conf/ijcnlp/LiSSLCN17} and DCR-Net~\cite{DBLP:conf/aaai/QinCLN020}, \ie 11,118 dialogues for training, 1,000 for validating, and 1,000 for testing.

The content of dialog in DailyDialog is based on human-designed scripts, specifically selected to elicit emotional expressions.
However, the human-written characteristic makes it hard to reflect actual emotional changes in real conversation scenes.
Besides, the average turn of DailyDialog is only 3.5, which means some dialogs may not have sufficient contextual information.
For this reason, we introduce our proprietary dataset named Customer Service Dialog.

\paratitle{Customer Service Dialog.} 
The dialogs of this dataset are collected from the ALIME\footnote{http://alixiaomi.com} platform, a robot platform for conversations with customers. The written language of this dataset is Chinese. Table~\ref{tab:basicstat} reports data characteristics of \datasetname. 
In this dataset, we also propose the notion of \textit{sentiment curve type}, which reflects the trend of emotion change in dialogs, \eg continuous type, improved type, falling type, U-shape type, and inverted U-shape type. The trend of emotion change could be an indicator of staff performance and user satisfaction.
Utterances and speaker types are both provided by ALIME. Intent, two-level emotions, user satisfaction, and emotion curve type are manually annotated afterwards.
We manually labeled $2,000$ dialogs, detailed in Table~\ref{tab:basicstat}.
Three annotators are hired to work independently for data annotation. The average of pair consistency exceeds 90\%, and the triple annotators' consistency reaches up to 85\%. The average pairwise kappa coefficients is 0.771. In short, we get high-quality annotations with the high level of consistency in labels.

\begin{table}
    \centering
    \caption{Data characteristics of the Customer Service dialog.}
    \vspace{0.7ex}
    \label{tab:basicstat}
    \begin{tabular}{l|r}
    \toprule
    Index    &  Value \\
    \midrule
    Average length of utterance & 9.4 \\
    Average turn of dialog & 16.1 \\  
    \midrule
    Number of Dialogs in Training Set & 1,400 \\
    Number of Dialogs in Validation set & 200 \\
    Number of Dialogs in Testing set & 400 \\
    \bottomrule
    \end{tabular}
\end{table}

Considering the task is to be conducted on Customer Service Dialog dataset, the set of utterance-level emotion $\mathcal{E}$ is \{\textit{Anger}, \textit{Dissatisfaction}, \textit{Worry}, \textit{Emotionlessness}, \textit{Happiness}, \textit{Comfort}\}. The set of speaker types is \{\textit{User}, \textit{Staff}\}. The categories of user satisfaction include \textit{Negative}, \textit{Neutral} and \textit{Positive}. The set of emotion curve categories is \{\textit{Concave}, \textit{Still}, \textit{Up}, \textit{Down}, \textit{Convex}\}.

\subsection{Implementation Details}

In our experiment, Glove~\cite{pennington2014glove} for English and Tencent AI Lab Embedding Corpus\footnote{https://ai.tencent.com/ailab/nlp/embedding.html}~\cite{song2018directional} for Chinese are used to initialize word vectors. The dimension of word vectors is 300 and 200, respectively.
The dimension of user embedding, staff embedding, intent embedding, and hidden vectors of $\RNN_u$ and $\RNN_d$ are set to 200.
There is a checkpoint every 16 mini-batch, and the batch size is 32. The dropout on embeddings is set to 0.5.

Adam~\cite{kingma2014adam} is used to optimize our model. The learning rate for model except word vectors are 1e-3, and 1e-4 for word vectors. In Adam, the two parameters $\beta_1$ and $\beta_2$ are 0.9 and 0.999. We use Pytorch\footnote{https://pytorch.org}~(version 1.4.0) to implement our model.

\begin{table*}
    \centering
    \caption{The macro precision, recall and $F1$ of emotion classification at utterance-level. The superscript $*$ of SVM and NB means the model uses uni-gram as feature. For superscript $\dag$, dense vectors gotten from the mean of word vectors are used as feature. The evaluation indicators marked with $^\#$ are reported in~\citet{DBLP:conf/aaai/QinCLN020}. The best results are in bold face and second best underlined. }

    \begin{tabular}{c|ccc|ccc}
    \toprule
    \multicolumn{1}{c|}{\multirow{2}[6]{*}{Model}} & \multicolumn{3}{c|}{\datasetname} & \multicolumn{3}{c}{DailyDialog} \\
\cmidrule{2-7}          & Precision &  Recall & F1 & Precision & Recall & F1\\
    \midrule
    SVM$^*$     &  43.4  & 28.8  & 30.1 & 31.8 & 17.1 & 17.8\\
    SVM$^\dag$  &  \underline{54.9}  & 22.9  & 23.3 & 21.7 & 15.8 & 15.7 \\
    NB$^*$      &  36.6  & \underline{47.2}  & 38.1 & 25.0 & 37.1 & 26.6\\
    NB$^\dag$   &  11.5  & 16.7  & 13.6 & 11.7 & 14.3 & 12.8\\
    LSTM        &  52.1 &	40.3 &  43.8 & \textbf{65.3} &	30.2 & 34.7\\
    AT-LSTM     &  52.4 &	40.3 &	43.9 & 60.4 &	30.2 & 34.4\\
    BERT-RNN    &  49.5 &  42.8 & 43.7 & 50.2 & 35.9 & 40.1\\
    DialogueRNN	& 47.0 & 44.4 & \underline{44.8} & 44.5$^\#$ &	37.7$^\#$ &	40.3$^\#$ \\
    DialogueGCN	& 49.3 & 42.9 & 42.4 & 56.1 & 31.7 & 35.2\\
    DCR-Net & 44.1 & 40.1 & 41.3 & 56.0$^\#$ &
    \underline{40.1}$^\#$ & \underline{45.4}$^\#$ \\
    \midrule
    Chat-Capsule    & \textbf{66.4} & \textbf{47.4} & \textbf{47.6} & \underline{61.5}	& \textbf{45.4} & \textbf{50.4}\\
    \bottomrule
    \end{tabular}
    
    \label{tab:utterance}
    \vspace{2ex}
\end{table*}

\subsection{Evaluation Details}

We conduct experiments to evaluate our model against baselines, including SVM, Naive Bayes~(NB), LSTM, AT-LSTM, BERT-RNN, DialogueRNN, DialogueGCN, and DCR-Net.
Note that, Chat-Capsule is not based on BERT. In order to observe the capabilities of our model, the comparison with BERT is deliberately added as well.

\paratitle{SVM and Naive Bayes.}
SVM~\cite{cortes1995support} utilizes maximum-margin hyperplane. Naive Bayes~(NB)~\cite{maron1961automatic} is based on Bayes' theorem with naive independence assumptions between features.
Features used for SVM and NB include: (i) uni-gram and (ii) the average of dense vector representations obtained through word2vec tools.

\paratitle{LSTM.} LSTM~\cite{hochreiter1997long} is a special kind of RNN, capable of learning long-term dependencies.
Bidirectional LSTM~(Bi-LSTM) is a variant of RNN, which is able to utilize the element's past and future information.

\paratitle{AT-LSTM.} AT-LSTM~\cite{wang2016attention} uses weight scores to attend the important part of input. To utilize the context information, bidirectional is used to utilize the element's past and future information.

\paratitle{BERT-RNN. }  BERT~\cite{DBLP:conf/naacl/DevlinCLT19}  is empirically powerful. We design BERT-RNN to encode dialog. BERT is used to encode each utterance, and RNN is used to encode the dialog, where the input is the utterance representations.

\paratitle{DialogueRNN. } DialogueRNN~\cite{DBLP:conf/aaai/MajumderPHMGC19} is based on recurrent neural network and keeps track of the individual party states throughout the conversation and uses this information for emotion classification.

\paratitle{DialogueGCN. } DialogueGCN~\cite{DBLP:conf/emnlp/GhosalMPCG19} based on graph neural network, leverages dependency of the interlocutors to model conversational context for emotion classification.

\paratitle{DCR-Net. } DCR-Net~\cite{DBLP:conf/aaai/QinCLN020}  considers the cross-impact and model the interaction between dialog act recognition and emotion classification by introducing a co-interactive relation layer.

\subsubsection{Emotion Evaluation at Utterance-level}

The distributions of emotions are imbalanced. However, emotions with small proportions, \eg ``anger'', ``dissatisfaction'' and ``comfort'' are considered more important. Therefore, we adopt macro precision, recall, and $F1$ as evaluation metrics.
Both DailyDialog~\footnote{http://yanran.li/dailydialog} and customer service dialog have utterance-level emotion, so we evaluate models on both datasets.

\paratitle{Observations.} Table~\ref{tab:utterance} reports the macro precision, recall and $F1$ of emotion classification at utterance-level. Neural network models perform better than traditional approaches as expected.
Attention based methods do not perform as expected due to: (i) the short length of utterance, and (ii) the best parameters in validation dataset performing not well in testing dataset.
BERT-RNN performs well on both \datasetname and DailyDialog datasets thanks to its powerful pre-training.
Respectively, DialogueRNN and DCR-Net achieve the second best results on \datasetname and DailyDialog due to their designs for conversation.
Chat-Capsule achieves the best performance on both datasets.
Interestingly, Chat-Capsule performs better than DCR-Net+BERT~(macro F1, 48.9 on DailyDialog dataset).
It is worth noting that a major limitation of DCR-Net is that it requires all utterances to predict utterance-level emotion, while Chat-Capsule only needs previous utterances to predict the current utterance's emotion.
Another limitation of DiagueRNN and DialgueGCN and DCR-Net is that they are unable to predict emotion at dialog-level.
Chat-Capsule is built to fully utilize the contextual information. Next, we show better performance of Chat-Capsule at dialog-level.

\subsubsection{Emotion Evaluation at Dialog-level}

We evaluate  emotion at dialog-level from two perspectives: (i) user satisfaction classification, and (ii) emotion curve type classification. The two perspectives can be useful for improving the quality of service for decision makers.

\begin{table}[]
    \centering
    \caption{User satisfaction evaluation on Customer Service Dialog. The superscript $^\#$ of DialogueRNN and DialogueGCN means the model is designed only for emotion classification at utterance-level.
    We use the average of the last three users' utterance-level emotion as the dialog-level user satisfaction. }
    \vspace{0.3ex}
    \begin{tabular}{l|cccc}
    \toprule

    Model     & Pre. & Rec. & F1 & Acc.\\
    \midrule
    SVM$^*$    &  52.2  &  45.4  &  44.9  &  69.3  \\
    SVM$^\dag$ &  38.1  &  34.4  &  28.5  &  64.0  \\
    NB$^*$     &  51.3  &  54.7  &  49.9  &  55.8  \\
    NB$^\dag$  &  21.2  &  33.3  &  25.9  &  63.5  \\
    LSTM       &  55.2 &	49.5 &	49.2 &	71.8  \\
    AT-LSTM    &  67.2 & \underline{55.0} &	\underline{57.6} &	\underline{73.0} \\
    BERT-RNN    & \textbf{71.9}	& 50.8 & 52.5 & 72.8\\
    DialogueRNN$^\#$ & 41.7 &	45.9 &	42.5 &	65.5 \\
    DialogueGCN$^\#$ & 40.0 &	41.6 &	38.3 &	65.0 \\

    \midrule
    Chat-Capsule     & \underline{71.3} & \textbf{64.8} & \textbf{67.2} & \textbf{77.0} \\
    \bottomrule
    \end{tabular}
    \label{tab:usersatisfaction}
    \vspace{1ex}
\end{table}

\paratitle{Observations.} Table~\ref{tab:usersatisfaction} reports the evaluation results for user satisfaction. Chat-Capsule achieves the best macro $F1$ of $67.2$.
As expected, neural network models perform better than  traditional machine learning models, \ie SVM, and NB with feature engineering.
Interestingly, we find the accuracy of SVM with uni-gram is higher than other baselines. It is caused by the imbalanced data, and this method tends to classify each dialog into ``neutral'' opinion.
The important part for emotion in this task of a dialog is the end, due to our motivation for evaluating the satisfaction of user in the whole session.
Emotion curve detection is reported in Table~\ref{tab:dia:curve}. Chat-Capsule is the best method, followed by LSTM, then traditional machine learning methods.
Interestingly, LSTM performs the second best. An aftermath shows that, due to data imbalance,  this method tends to classify every dialog into ``Still'' curve.
Similar to user satisfaction, the macro $F1$ is more meaningful than accuracy.

\subsection{Ablation study}

We study the effect of the rectifier module and feedback module in utterance-level capsules. The macro F1 of Chat-Capsule without rectifier module is 46.4, which is 1.2 points lower than Chat-Capsule. The macro F1 of Chat-Capsule without both rectifier module and feedback module is 43.2, which is 4.4 points lower than Chat-Capsule. The ablation study suggests that the carefully designed modules in capsules are able to improve the representation capability of the model.

\begin{table}[]
    \centering
    \caption{The emotion curve evaluation of \datasetname. }
    \vspace{0.7ex}
    \begin{tabular}{l|cccc}
    \toprule

    Model     & Pre. & Rec. & F1 & Acc.\\
    \midrule
    SVM$^*$    &  22.1   &  23.3   &  22.6  &  52.5   \\
    SVM$^\dag$ &  17.0  &  20.0  &  15.5  &  59.5   \\
    NB$^*$     &  \underline{26.0}  &  26.8  &  25.1   &  42.0  \\
    NB$^\dag$  &  12.0  &  20.0  &  15.0  & 60.0 \\
    LSTM      & 25.7 &	27.2 &	\underline{25.8} &	\underline{60.8} \\
    AT-LSTM    & 23.7 &	\underline{27.7} &	25.6 &	55.8 \\
    BERT-RNN    & 25.4	& 26.1 & 25.3 & 58.8\\
    \midrule
    Chat-Capsule & \textbf{42.4} & \textbf{32.3} & \textbf{33.1} & \textbf{63.3}\\
    \bottomrule
    \end{tabular}
    \label{tab:dia:curve}
    \vspace{1ex}
\end{table}

\section{Conclusion}
\label{sec:conclusion}

In this paper, we study  dialog-level emotion analysis. We observe that intent and speaker type of utterance affect its emotion. More importantly, emotion is contextually related. Motivated by these observations, we propose Context-based Hierarchical Attention Capsule~(Chat-Capsule) for dialog-level emotion analysis.
Chat-Capsule is capable of not only attending to the contextual information, but also considering the intent and speaker type.
Experiments show that our proposed Chat-Capsule achieves start-of-the-art performance.

\bibliography{anthology,custom}
\bibliographystyle{acl_natbib}

\end{document}